\documentclass[runningheads,a4paper]{llncs}
\usepackage[T1]{fontenc}
\usepackage[utf8]{inputenc}
\usepackage{lmodern}
\usepackage{pbox}
\usepackage{rotating}
\let\llncssubparagraph\subparagraph
\let\subparagraph\paragraph
\usepackage{titlesec}
\let\subparagraph\llncssubparagraph

\usepackage[pdftex,bookmarks,colorlinks=false]{hyperref}

\usepackage{amssymb} 
\usepackage{graphicx}
\usepackage{url}
\usepackage{float}
\usepackage{xspace}
\newcommand{\keywords}[1]{\par\addvspace\baselineskip
\noindent\keywordname\enspace\ignorespaces#1}

\newcommand{\strutUp}{\rule{0pt}{3ex}}

\newcommand{\df}{\ensuremath{\textrm{\bf df}}\xspace}
\newcommand{\cf}{\ensuremath{\textrm{\bf cf}}\xspace}
\begin{document}

\mainmatter

 \titlerunning{Unsupervised Keyword Extraction from Polish Legal Texts}

 \authorrunning{Unsupervised Keyword Extraction from Polish Legal Texts}

 \author{Michał Jungiewicz\inst{1,2} \and Michał Łopuszyński\inst{1}}

 \title{Unsupervised Keyword Extraction From Polish Legal Texts}
 \institute{
 Interdisciplinary Centre for Mathematical and Computational Modelling,\\
 University of Warsaw,
 Pawińskiego 5a, 02-106 Warsaw  Poland, \\
 \email{m.lopuszynski@icm.edu.pl} \\
 \and
 Faculty of Electronics and Information Technology, \\
 Warsaw University of Technology,
 Nowowiejska 15/19, 00-665 Warsaw, Poland}

 \maketitle

 \begin{abstract}
 In this work, we present an application of the recently proposed
 unsupervised keyword extraction algorithm RAKE to
 a corpus of Polish legal texts from the field of  public procurement.
 RAKE is essentially a language and domain independent method.
 Its only language-specific input is a stoplist containing a set of
 non-content words. The performance of the method heavily depends
 on the choice of such a stoplist, which should be domain adopted.
 Therefore, we complement RAKE algorithm with an automatic
 approach to selecting non-content words, which is based
 on the statistical properties of term distribution.

\bigskip

\begin{scriptsize}
This paper was published in
``Advances in Natural Language Processing,
  9th International Conference on NLP, PolTAL 2014, Warsaw, Poland, September 17-19, 2014.
  Proceedings'', Lecture Notes in Computer Science, Volume 8686.
  The final publication is available at \href{http://dx.doi.org/10.1007/978-3-319-10888-9_7}{link.springer.com}.
\end{scriptsize}
 \end{abstract}

\section{Introduction}

Automatic analysis of legal texts is currently viewed as a promising
research and application area~\cite{Francesconi2010}.  On the other hand,
keyword extraction is a very useful technique in organization of large
collections of documents.  It helps to present the available information to
the user, aids browsing and searching.  Moreover, extracted keywords can be
useful as features in tasks, such as document similarity calculation,
clustering, topic modelling, etc.

Unfortunately, the problem of automatic keyword extraction is far from
solved.  A recently conducted competition during the SemEval 2010 Workshop,
showed that the best available algorithms do not exceed 30\% of the
F-measure, on the manually labeled test documents~\cite{Kim2010}. It is worth
noticing that these tests were based on English texts. For highly inflected
languages (e.g., Polish) it might be even more difficult and algorithms here
are certainly less developed and verified.

In the presented paper, we employ recently proposed RAKE
algorithm~\cite{Rose2010}.  It was designed as an unsupervised,
domain-independent, and language-independent method of extracting keywords
from individual documents. These features make it a promising candidate
tool for a highly specific task of extracting keywords from Polish legal
texts.  However, in the original paper authors evaluated RAKE only on
English texts. Its performance on a very different Slavic
language may deviate and is worth verifying.

The corpus used in this research consisted of 11 thousand rulings of the
National Appeals Chamber from the Polish Public Procurement Office.  In our
opinion, this set of documents is particularly interesting and challenging.  It
contains very diverse vocabulary, not only related to law and public
procurement issues, but also to the technicalities of discussed contracts
coming form very different fields (medicine, construction, IT, etc.)

\section{Automatic Stoplist Generation \label{sec:AutomaticStoplistGeneration}}
The general idea behind RAKE algorithm is based on splitting a given text
into word groups isolated by sentence separators or words from a provided
stoplist. Each such a word group is considered to be a keyword candidate
and is scored according to the word co-occurrence graph. The details of the
method can be found in~\cite{Rose2010}.  The stoplist constitutes the most
important
``free parameter'' of RAKE, as it is the only way to adjust this algorithm
to the specific language and domain.  As recognized by the authors of RAKE,
it is also a crucial ingredient on which the effectiveness of the algorithm
strongly depends~\cite{Rose2010}. Our initial tests carried out with a
standard information retrieval stoplist yielded poor results for the case
of Polish legal texts.  There were a lot of very long keywords, containing
many uninformative words, even though our implementation did not include
merging of the adjoining keyword candidates. Sample results are presented
in Table~\ref{tab:KeywordSummary}A.  To alleviate this type of problems,
the authors of RAKE propose two methods of automatic stopwords generation
from a given corpus~\cite{Rose2010}.  However, none seems satisfactory for
us. The first one is very crude, as it simply uses the most frequent
words. The second one requires an annotated training set (supervised
learning).  Therefore, we develop our own unsupervised approach to the
stoplist auto-generation problem.  It is based on the observation that
distribution of the number of occurrences per document for stopwords
usually follows typical random variable model (e.g., Poisson
distribution). Informative content words, on the other hand, occur in more
``clustered'' fashion and mostly deviate from the distribution of
stopwords~\cite{Church1995,Manning1999}.
\begin{sidewaysfigure}
\includegraphics[width=0.49\textwidth]{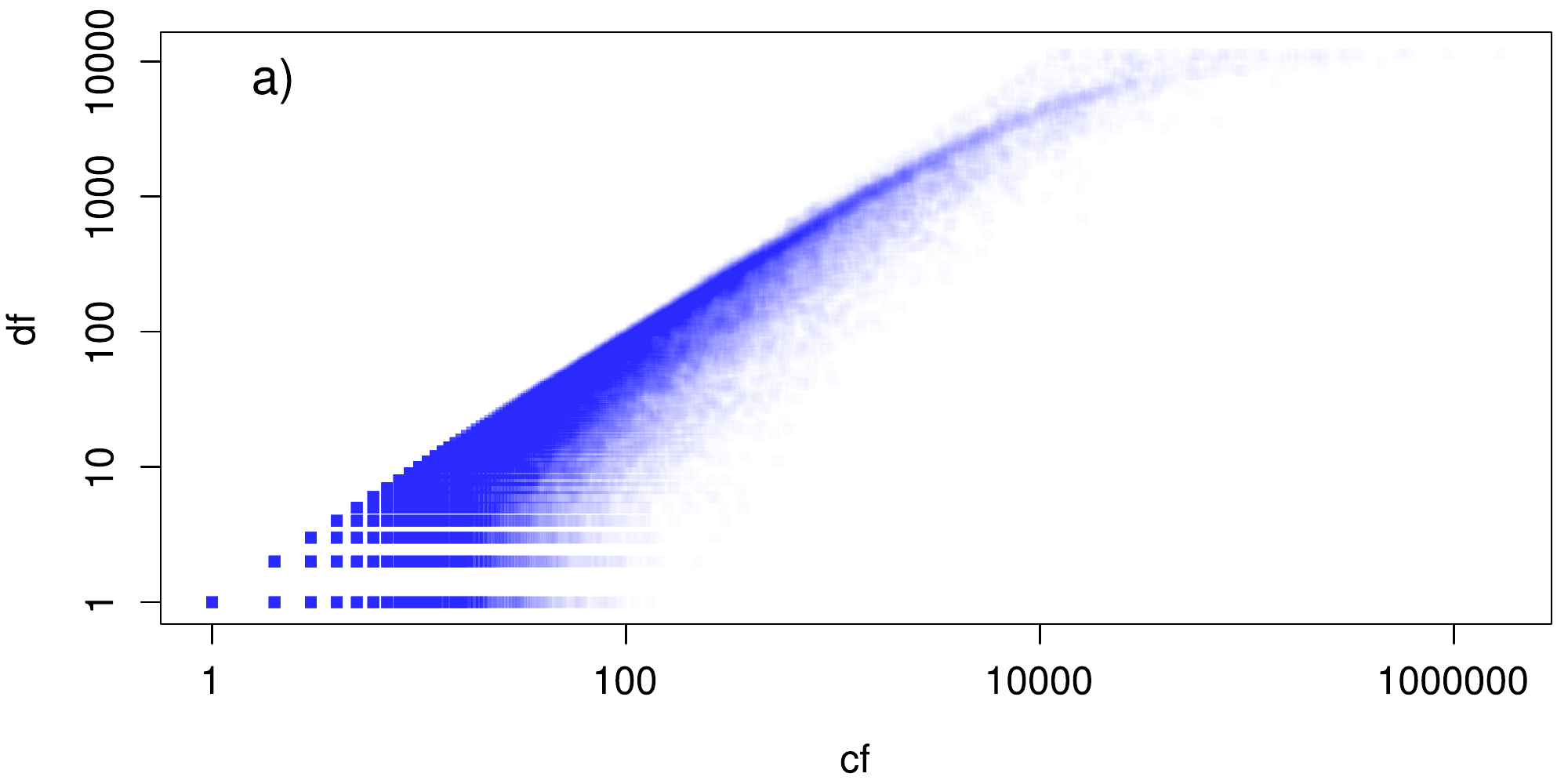}
\includegraphics[width=0.49\textwidth]{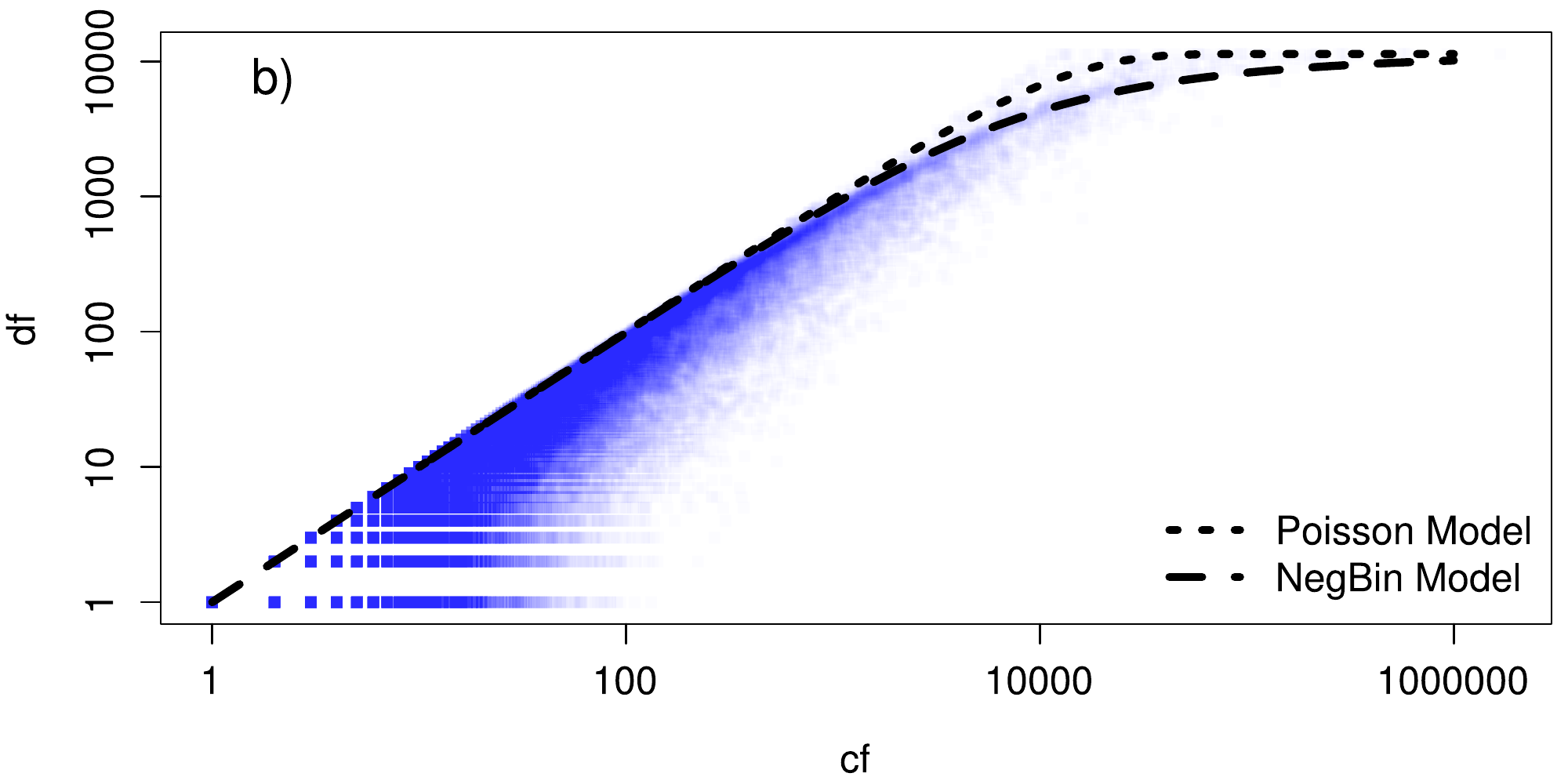}

\includegraphics[width=0.49\textwidth]{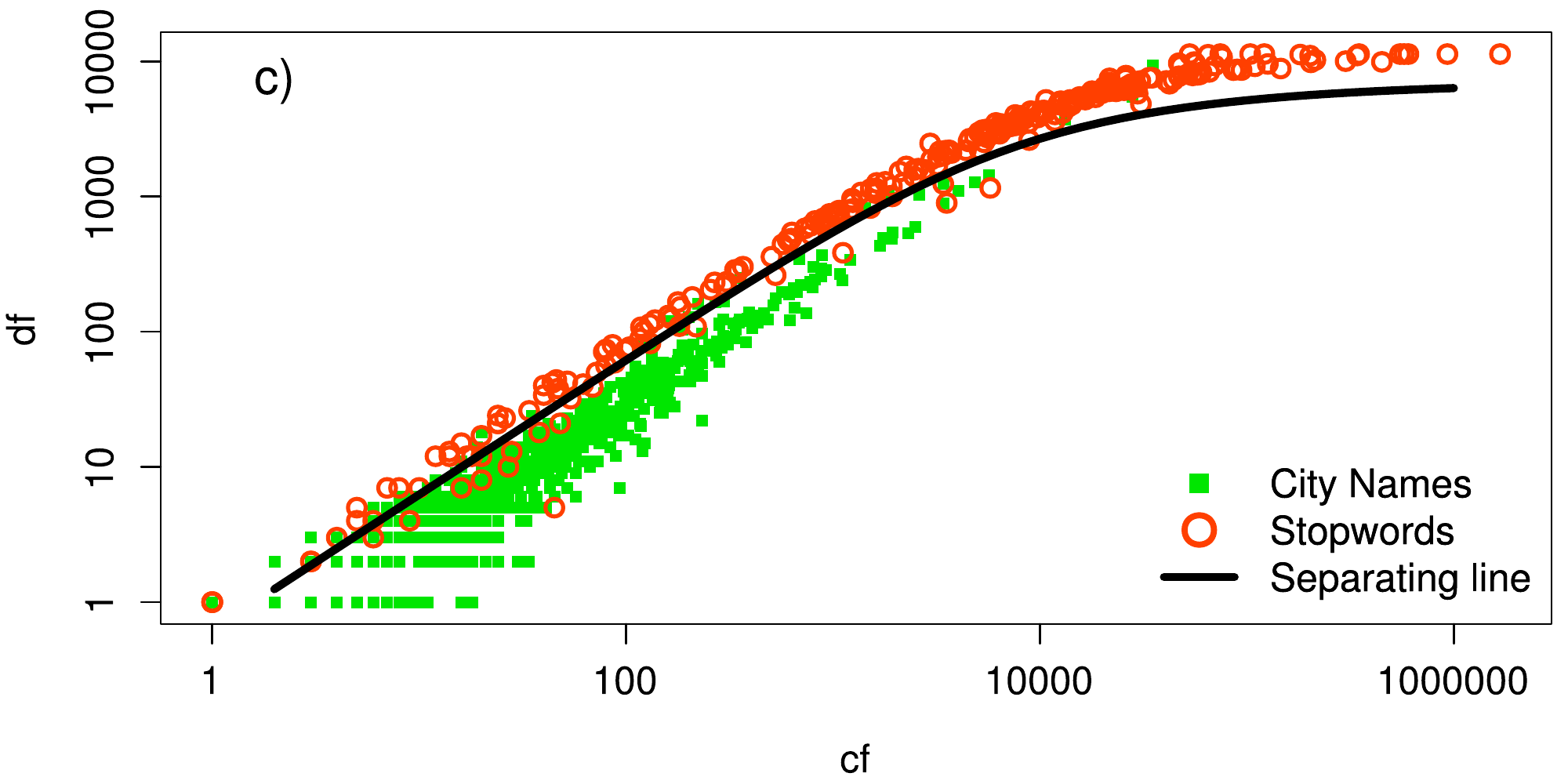}
\includegraphics[width=0.49\textwidth]{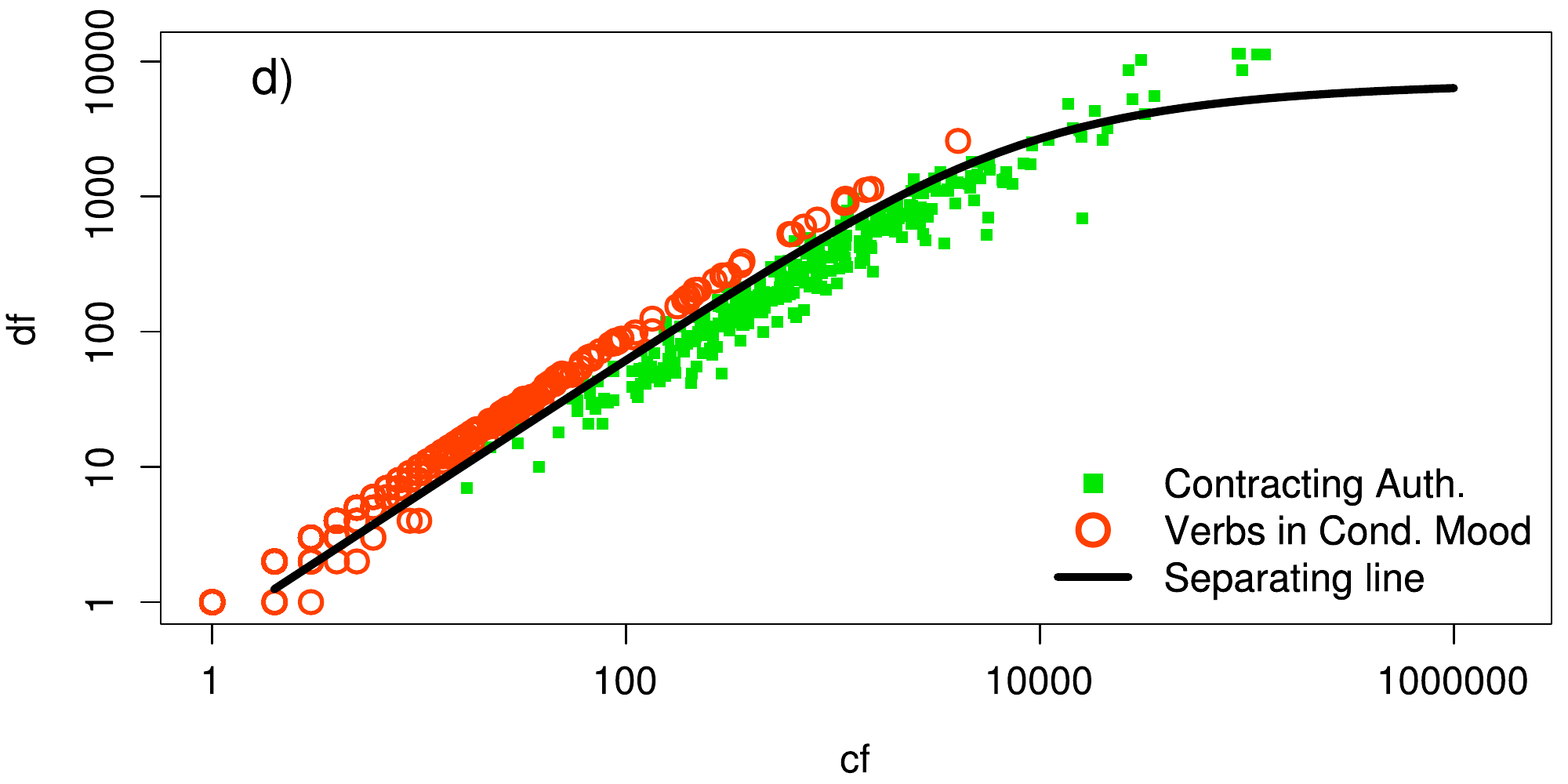}
\caption{Scatter plots of the number of documents with a given
         word ({\bf df}) vs. its frequency
         in the collection ({\bf cf}) for the whole corpus vocabulary.
         Panel a) shows plain scatter plot.
         Panel b) compares the model making use of the Poisson distribution
         (\ref{eq:PoissonModel}) with the negative binomial approach
         (\ref{eq:NegBinModel}) \label{fig:ScatterPlot}.
         Panel c)
         examines the location of the city names and
         standard information retrieval stopwords.
         Panel d) contrasts the location of contracting authorities and verbs in
         conditional mood.  Clearly, non-content words
         (information retrieval stopwords and verbs in conditional mood)
         tend to locate
         close to the theoretical curve given by
         (\ref{eq:NegBinModel}).  We decided to extract the
         non-content words for stoplist using the separating line
         $\overline{\df}/\df = 1.6$, which is marked in panels c) and d)
         \label{fig:ScatterPlot}}
\end{sidewaysfigure}

The simplest method of detecting this deviation is based on two variables
--- the number of documents in which a given word is present \df and the
cumulative collection word frequency \cf.  For the randomly distributed
stopwords the relation of \df to \cf in a large set of documents is defined
by the probability theory~\cite{Manning1999}
\begin{equation}
\label{eq:GeneralModel}
\overline{\df}\;(\cf)= N (1 - P (0,\mu=\cf/N)),
\end{equation}
where $N$ is the total number of documents, and $P(0,\mu)$ is
the probability of the word occurring 0 times, provided its average number
of occurrences per document $\mu$ (by definition $\mu=\textrm{\bf cf}/N$).
For the simplest Poisson model the equation reduces to
\begin{equation}
  \overline{\df}\;(\cf) = N (1 - \exp(-\cf/N)).
  \label{eq:PoissonModel}
\end{equation}
The plot of \df against \cf for all words in
the examined corpus is presented in Fig.~\ref{fig:ScatterPlot}a.
The Poisson model is plotted in Fig.~\ref{fig:ScatterPlot}b.
One can easily see that it does not give an accurate description
for the high values of \cf.
Therefore, we decided to replace the Poisson distribution with
the negative binomial model. It is closely related to
the Poisson variable, but allows for larger variance.
It can be also represented as infinite combination of Poisson
distributions with different~$\mu$.
After substituting the negative binomial probability distribution
function for $P(0,\mu)$ in (\ref{eq:GeneralModel}), we get
\begin{equation}
\overline{\df}\;(\cf) =
   N \left( 1 - \left ( 1 + \frac{\cf}{Nr} \right )^{-r} \right ),
\label{eq:NegBinModel}
\end{equation}
where $r > 0$ is the additional parameter of the negative
binomial distribution. In the case of $r \to \infty$ with fixed
$\mu$, the negative binomial variable converges to
the Poisson model. In Fig.~\ref{fig:ScatterPlot}b, we compare the
predictions of~(\ref{eq:PoissonModel})~and~(\ref{eq:NegBinModel})
with the value of $r=0.42$, adjusted to fit the data.
It is easily seen that the description of the high \cf region improves
for the negative binomial case.

To further illustrate the difference between the content and non-content
words, we compared locations of a few sample word categories in
(\cf,\df) space. We selected two groups of non-informative words, namely,
the usual information retrieval stopwords (containing conjunctions,
pronouns, particles, auxiliary verbs, etc.) and a class of verb forms in
conditional mood, ending on \emph{-łaby} \emph{-łoby}. These two groups
were compared with two categories of words which definitely carry
important information, i.e., the names of the cities and the most frequent
words extracted from the contracting authorities list (cleaned from
stopwords and city names to avoid overlapping categories). The comparison
is presented in Fig.~\ref{fig:ScatterPlot}c and ~\ref{fig:ScatterPlot}d.  The
displayed graphs confirm the assumption of larger deviation from the
negative binomial distribution in the case of content words. Approximate
separation can be obtained by $\overline{\df}/\df<1.6$.  The terms
satisfying this condition and occurring in more than ten documents were
used as stoplist in RAKE keyword extraction algorithm later on.

\section{Preliminary Results \label{sec:PreliminaryResults}}

\begin{sidewaystable}
\setlength{\tabcolsep}{8pt}
\caption{Summary of experiments with RAKE. Both the original keywords
         and their English translations are given
         \label{tab:KeywordSummary}}
 %Both the extracted keywords and their translations are given.
 %Results from part A were obtained with a standard
 %information retrieval stoplist,
 %whereas parts
 %B--D were produced with the stoplist generated using
 %method from
 %Sect.~\ref{sec:AutomaticStoplistGeneration}.
 \vspace{0.3cm}
\centering
\begin{tabular}{p{0.45\textwidth}p{0.45\textwidth}}
\hline\noalign{\smallskip}
\multicolumn{2}{l}{{\bf A.}
 {\bf Top 5 high-score keywords extracted from a sample document
      (standard stoplist)}} \\
\hline\noalign{\smallskip}
samej grupy kapitałowej dotyczącego wykonawcy Przedsiębiorstwo
Usług Komunalnych Empol sp.
& the same capital group concerning the contractor
  Municipal Services Company Empol \\
\strutUp Dzienniku Urzędowym Unii Europejskiej 23 marca 2013 r.
&  (in) the Official Journal of the European Union 23 March 2013 \\
\strutUp Prezesa Krajowej Izby Odwoławczej 20 czerwca 2012 r.
& Chairman of the National Appeal Chamber 20 June 2012 \\
\strutUp pierwszej kolejności Krajowa Izba Odwoławcza winna ocenić &
firstly the National Appeal Chamber should judge \\
\strutUp Krajowa Izba Odwoławcza uwzględniła odwołanie
konsorcjum Sita Małopolska &
National Appeal Chamber has upheld the appeal of the Sita Małopolska
Consortium \\
\hline\noalign{\smallskip}
\multicolumn{2}{l}{{\bf B.}
 {\bf All keywords extracted from a sample document
  (auto-generated stoplist of Sect. \ref{sec:AutomaticStoplistGeneration})}} \\
\hline\noalign{\smallskip}
Przedsiębiorstwo Usług Komunalnych Empol & Municipal Services Company Empol \\
przedsiębiorstwo usług komunalnych & municipal services company \\
zagospodarowanie odpadów komunalnych & management of municipal waste \\
odbieranie odpadów komunalnych &  municipal waste collection \\
właścicieli nieruchomości zamieszkałych & residential real estate owner \\
konsorcjum Sita Małopolska & Consortium Sita Małopolska \\
Sita Małopolska & Sita Małopolska \\
grupy kapitałowej & capital group \\
 \hline\noalign{\smallskip}
\multicolumn{2}{l}{
 {\bf C.} {\bf Most frequent keywords in the whole corpus
   (auto-generated stoplist of Sect. \ref{sec:AutomaticStoplistGeneration})}} \\
\hline%\noalign{\smallskip}
roboty budowlane   & construction works   \\
robót budowlanych   & construction works (different form) \\
konsorcjum firm    & consortium of companies \\
ograniczoną~odpowiedzialnością \hspace{1cm} & limited liability \\
formularzu ofertowym & offer form \\
\hline\noalign{\smallskip}
\multicolumn{2}{l}{ {\bf D.} {\bf Most frequent keywords with four tokens
 (auto-generated stoplist of Sect. \ref{sec:AutomaticStoplistGeneration})}} \\
\hline\noalign{\smallskip}
PKP Polskie Linie Kolejowe & PKP Polish State Railways \\
Generalnej Dyrekcji Dróg Krajowych  &
   General Directorate for National Roads \\
samodzielny publiczny szpital kliniczny &
   independent public clinical hospital \\
wykazu wykonanych robót budowlanych  &
   list of conducted construction works\\
GE Medical Systems Polska
   & GE Medical Systems Poland \\
\hline
\end{tabular}
\end{sidewaystable}

After developing the method of automatically distilling the stopword list
from a given corpus, we ran the keyword extraction procedure on the available
documents.  Since the documents did not contain any manually assigned
keywords, we can do only qualitative analysis at this stage.  The
preliminary results are presented below.

We found that the method indeed yields useful key phrases.
Its results for a sample document are presented in
Table~\ref{tab:KeywordSummary}B and can be compared
with the results obtained using standard information retrieval stoplist
(Table~\ref{tab:KeywordSummary}A).
The extracted phrases look promising, as they clearly indicate
the topic of municipal waste management to which the analyzed document
is related.

To get more insight into the behaviour of the algorithm throughout the
whole corpus, we also analyzed the most frequently detected keywords.
Top five most popular key phrases are presented in
Table~\ref{tab:KeywordSummary}C.
The result is intuitively well understood, since a considerable part of
the public procurement contracts in Poland (in the period 2007--2013,
covered by the analyzed corpus) deals with large scale construction
works carried out by consortia consisting of a few companies.
This is clearly reflected in the obtained results.

Obviously, the most frequently occurring keywords from
Table~\ref{tab:KeywordSummary}C are rather general. However, if we restrict
ourselves to longer phrases, we can easily check that their vagueness
decreases and that they still form meaningful and informative word groups.
Analyzing the most popular four token key phrases
(Table~\ref{tab:KeywordSummary}D), we found that RAKE method is capable
of extracting names of large contracting authorities and companies.  This
also seems a very desirable behaviour of the algorithm.
Of course, in order to quantify the performance of the algorithm,
rigorous tests based on the human expert knowledge are necessary.

\section{Summary and Outlook \label{sec:Summary}}
In this paper, we have presented a work in progress report on the
unsupervised keyword extraction from Polish legal texts. We have employed
recently proposed RAKE algorithm and extended it with the automatic, corpus
adopted stoplist generation procedure. Qualitative tests of the method
indicate that the approach is promising. In the future, we plan
quantitative tests, however, this has to involve human domain
experts and hence is a lengthy process.  In addition, we plan also
further optimization of the method. Introducing stemming and adjusting
keyword ranking scheme of RAKE algorithm seem to be the most attractive
directions.

\subsubsection*{Acknowledgments.} We acknowledge the use of computing
facilities of the Interdisciplinary Centre for Mathematical and Computational
Modelling within the grant G57-14.

\bibliographystyle{unsrt}
\bibliography{biblio}

\begin{thebibliography}{1}

\bibitem{Francesconi2010}
Enrico Francesconi, Simonetta Montemagni, Wim Peters, and Daniela Tiscornia.
\newblock {\em Semantic Processing of Legal Texts}.
\newblock Springer, Berlin; New York, 2010.

\bibitem{Kim2010}
Su~Nam Kim, Olena Medelyan, Min-Yen Kan, and Timothy Baldwin.
\newblock {SemEval-2010} task 5: Automatic keyphrase extraction from scientific
  articles.
\newblock In {\em Proceedings of the 5th International Workshop on Semantic
  Evaluation}, SemEval '10, page~21, Stroudsburg, PA, USA, 2010. Association
  for Computational Linguistics.

\bibitem{Rose2010}
Stuart Rose, Dave Engel, Nick Cramer, and Wendy Cowley.
\newblock Automatic keyword extraction from individual documents.
\newblock In Michael~W. Berry and Jacob Kogan, editors, {\em Text Mining.
  Applications and Theory}, page~1. John Wiley and Sons, Ltd, 2010.

\bibitem{Church1995}
Kenneth~W. Church and William~A. Gale.
\newblock Poisson mixtures.
\newblock {\em Natural Language Engineering}, 1(02):163, 6 1995.

\bibitem{Manning1999}
Christopher~D. Manning and Hinrich Sch\"utze.
\newblock {\em Foundations of statistical natural language processing}.
\newblock MIT Press, Cambridge, Mass., 1999.

\end{thebibliography}
\end{document}